\pdfoutput=1

\documentclass{svproc}

\usepackage{amsmath}
\usepackage{amsfonts}

\usepackage{algorithm}
\usepackage{algpseudocode}

\usepackage[dvipdfm]{graphicx} 

\usepackage[square,numbers]{natbib}   
\usepackage{cite}

\usepackage{url}

\usepackage{xcolor}

\begin{document}
\mainmatter              
\title{Grouping Shapley Value Feature Importances of Random Forests for explainable Yield Prediction}
\titlerunning{Grouping Shapley Values for Yield Predictions}  
%
\author{Florian Huber$^\textrm{1*}$, Hannes Engler$^\textrm{2}$, Anna Kicherer$^\textrm{2}$, Katja Herzog$^\textrm{2}$, Reinhard Töpfer$^\textrm{2}$ and Volker Steinhage$^\textrm{1}$}
\authorrunning{Florian Huber et al.} 
%
%
\institute{$^\textrm{1}$Department of Computer Science IV, University of Bonn, Friedrich-Hirzebruch-Allee 8, 53121 Bonn, Germany\\
$^\textrm{2}$
Julius Kühn-Institut, Federal Research Centre of Cultivated Plants, Institute for Grapevine Breeding Geilweilerhof, 76833 Siebeldingen, Germany \\
\email{$^\textrm{*}$huber@cs.uni-bonn.de}}

\maketitle              

\begin{abstract}
Explainability in yield prediction helps us fully explore the potential of machine learning models that are already able to achieve high accuracy for a variety of yield prediction scenarios. The data included for the prediction of yields are intricate and the models are often difficult to understand. However, understanding the models can be simplified by using natural groupings of the input features. Grouping can be achieved, for example, by the time the features are captured or by the sensor used to do so. The state-of-the-art for interpreting machine learning models is currently defined by the game-theoretic approach of Shapley values. To handle groups of features, the calculated Shapley values are typically added together, ignoring the theoretical limitations of this approach. We explain the concept of Shapley values directly computed for predefined groups of features and introduce an algorithm to compute them efficiently on tree structures. We provide a blueprint for designing swarm plots that combine many local explanations for global understanding. Extensive evaluation of two different yield prediction problems shows the worth of our approach and demonstrates how we can enable a better understanding of yield prediction models in the future, ultimately leading to mutual enrichment of research and application. 

\end{abstract}
%
%
\section*{\begin{normalsize}\textit{Preprint accepted at IntelliSys 2023}\end{normalsize}}
\noindent
$\copyright$ 2023. This manuscript version is made available under the CC-BY-NC-ND 4.0
license https://creativecommons.org/licenses/by-nc-nd/4.0/

\section{Introduction}

Providing a secure source of food for the world's population is an increasingly difficult challenge. While the world population grows, the available farmland around the world is limited, pushing us to reach the limits of efficiency in agricultural environments. One way to ensure precise planning and optimized workflows is to predict yields with machine learning approaches, which have developed increasingly high accuracy over time \citep{van2020crop}. However, with increasing accuracy, the complexity of the models and input data also expands. Popular data sources include remote sensing data \citep{huber2022extreme}, fertilization schedules \citep{meng2021predicting}, climate information \citep{cao2020identifying}, and soil evaluation \citep{sirsat2019machine}, just to name a few. Successful models utilize up to over 1000 features to predict yield \citep{huber2022extreme}, making models and data increasingly difficult to understand.

The investigation of yield prediction models is a time-consuming part of the research. Often, models are investigated by defining feature subsets and correlating different accuracies reached, when these subsets are absent from training, with their importance for yield prediction \citep{you2017deep, bobeda2018using, cao2020identifying}. Not only is the number of possible feature combinations exponential in the number of groups to be investigated, full training, tuning, and testing steps are also necessary for every combination. \citet{lundberg2017unified} argue that Shapley values are capable of offering a unified framework for explainability and provide a fast way to calculate Shapley value feature importances for tree structures \citep{lundberg2020local}. While Shapley value feature importances offer a sound mathematical foundation for desirable properties, like the individual feature explanations adding up to the prediction output of the models, they are not well suited to investigate models relying on a high amount of input features and, therefore, are not naturally equipped to explain yield prediction models. Without grouping features together, it is harder to find patterns within the data or even present them in a meaningful way. Furthermore, when using feature importances to decide which sensors to buy, or at which times a plant should be monitored, it is mostly a batch of data that is procured together. Grouping features for feature importances allows us to evaluate all connected data as a unit. 

The aim of our study is to address these problems and provide a global explanation of yield prediction models through a direct extension of Shapley values. We want to preserve the desirable mathematical properties and stay in line with the unified approach, while extending the paradigm to be able to handle large groups of features. Yield prediction tasks usually have some properties in common that we can exploit to do so. First, within the input data, we can pick out similar features and group them to obtain explanations. A multitude of features are often captured by the same sensors or associated with a similar time frame. Second, tree structures achieve state-of-the-art results for yield prediction in many use cases \citep{huber2022extreme, cao2020identifying, sirsat2019machine}. Combining this, we are able to focus on feature groups to calculate Shapley value feature importances. Furthermore, focusing on tree structures allows for fast polynomial-time calculation. We will refer to this paradigm as \textbf{G}rouped \textbf{S}hapley \textbf{V}alues (GSV). Shapley value feature importances are local explanations, meaning that they are calculated for individual data instances. Within this study, our further objective is to take advantage of many local explanations to obtain information on the global structure of the model by designing specific swarm plots. 

Achieving accessible explanations for tasks with a multitude of input features will help the exchange of information between computer scientists and domain experts, which is very important to model the intricacies for the prediction of different yields. Furthermore, being able to explain why the model made its prediction can help to accept automated yield prediction within the society. The information gained can be used to decide which sensors are worth buying and which plant steps are worth monitoring for future yield prediction surveys. 

In summary, the contributions of this study are the following:
\begin{enumerate}
    \item We examine a definition of GSV for predefined coalitions that conserves the axiom of efficiency, which is important for explainable machine learning applications (cf. Section \ref{sec:GSP}).
    \item We describe a graphical structure that allows the combination of many local explanations to unlock global understanding of the models (cf. Section \ref{sec:ExplainByGSV}).
    \item To our knowledge, we are the first to describe a polynomial algorithm to calculate GSV feature importances for random forests, based on a definition where groups of features compete directly against other groups (cf. Section \ref{sec:algo}).
    \item We show the capabilities of our approach by providing an exemplary analysis of two different yield prediction models (cf. Section \ref{sec:exp}).
\end{enumerate}

The rest of the document will be presented as follows: We summarize the related articles that influence our research in Section \ref{sec:related_work}. In Section \ref{sec:Methodology} we describe our approach to GSV, give an idea of the polynomial-time algorithm to calculate the GSV on tree structures (Algorithm \ref{alg:groupshap}), and explain how we leverage many local explanations to obtain global model understanding. An example of the usage of our algorithm in the context of yield prediction is presented in Section \ref{sec:exp} and we draw conclusions of our work in Section \ref{sec:Conclusion}. A detailed presentation of Algorithm \ref{alg:groupshap} is given in the Appendix.
\section{Related Work}
\label{sec:related_work}
We review related work categorized. First, we will examine the state of explainability in yield prediction. Second, we will describe the history of using Shapley values as feature importance measure together with previous efforts to expand the paradigms of Shapley values onto groups of features, both in a general game-theoretic and a machine learning context. 


Yield prediction as a research problem is being addressed with a multitude of approaches, both for modeling and explaining the results. Many deep learning approaches are applied to yield prediction; for example, the work of \citet{you2017deep}, \citet{wang2020winter}, and \citet{khaki2021simultaneous}, just to name a few. Although the problem is difficult due to the black-box nature of deep learning models, some efforts were made to explain the results. \citet{you2017deep} correlate the importance of a feature with the decrease in the accuracy of the model when information on the feature is missing. They render the information of a feature useless by randomly permuting the values throughout the dataset. With this approach, they can assess the importance of whole feature groups by permuting their values simultaneously. Experiments show that the red and near-infrared bands of satellite images are important for their yield prediction model. Another way to gain explanations for the output of deep learning frameworks is deployed by \citet{khaki2021simultaneous}. They backpropagate the output of active neurons in the last layer and are able to find active neurons in the first layer to correspond to features of the input space. 

An explainable alternative to deep learning for yield prediction can be found using regression forests. Tree structures are prevalent in yield prediction and show state-of-the-art results in a multitude of real-world scenarios. \citet{huber2022extreme} have derived state-of-the-art results for the prediction of soybean yields in the United States. They applied regression forests created by E\textbf{x}treme \textbf{G}radient \textbf{Boost}ing (XGBoost) \citep{chen2016xgboost}. Feature importances are analyzed by adding multiple Shapley value feature importances over the whole training data to find that the red and near-infrared bands are very important, especially in the time close to harvest. \citet{Diaz_2017} use the M5-Prime algorithm \citep{wang1996induction} to create regression trees for the prediction of citrus orchards in Argentina. Since singular trees are considered instead of forests, they conduct a visual analysis of the feature importances based on the resulting tree structure. Similarly, \citet{bobeda2018using} use the M5-prime algorithm to predict citrus orchards in Argentina. They use another popular method to understand the yield predictions and explain their model output. By creating multiple subsets of features and evaluating their model in the absence of each of the subsets, they find that it is not necessary to count the fruits in the trees multiple times a year, and the results are only slightly worse when, instead, calculating the volume of the trees' crowns once. 

Since we can understand how tree models are functioning, we can rely on the inner relations of the trees to find feature importances that can give explanations to the model's output. Most famously, the \textbf{M}ean \textbf{D}ecrease in \textbf{I}mpurity (MDI) \citep{louppe2013understanding} can be used to give a measure of the number of splits made by each feature, weighted with the impact of the individual split, that is, the proportion of training samples divided. This internal measure of importance is used by \citet{sirsat2019machine} to select expressive features when predicting grapevine yields based on phenological information, soil properties, and climatic conditions. \citet{meng2021predicting} use this method to show the high importance of vegetation indices when predicting maize yield in California on a field scale.



Shapley values are a game-theoretic measure for solving a fair distribution of resources in a cooperative game. The value awarded to a player is calculated by averaging his contribution over all possible coalitions that he could join within the game. The Shapley value was first used as a feature importance measure by \citet{lundberg2017unified}. The idea is to assign each feature an importance value for a particular prediction. The choice of the game-theoretic construct to solve cooperative games, namely the Shapley value, was made because of its desirable theoretical properties and results that are in line with human intuition. One of the mathematical properties allows them to provide an additive feature attribution method, which means that the sum of the feature importances will equal the actual model output for the example. 

The calculation of Shapley values is, in general, a NP-hard problem. But for decision trees, exploiting the tree structure allows computations in polynomial time, as explained in \citep{lundberg2020local}, giving the first polynomial-time algorithm to compute explanations on tree structures based on game theory. The work also gives an idea of how to use many local explanations to represent the global structure of the model. 

Lastly, we want to highlight other efforts to extend the Shapley value feature importances towards groups of features. On the one hand, we have the classical game-theoretic view on this topic. However, the relevant works \citep{grabisch1999axiomatic, marichal2007axiomatic, flores2019evaluating} all fail to preserve the efficiency property, which means that the sum of all attribution values of the features will not coincide with the output of the model and, therefore, are not suitable to base the explanations on. The work of \citet{jullum2021groupshapley} recognizes this weakness and presents a form for extending the Shapley value to groups of players in the context of feature importance. This allows for easier representation of the results together with a lower computational complexity. \citet{amoukou2021shapley} base an approach to evaluate groups of features on a different definition of grouped Shapley values, where groups of players continue to play against individuals. The work is extended by giving a fast computation for tree structures and selecting minimal subsets of features, so that the classifier will make the same decision with high probability. 


\section{Approach}
\label{sec:Methodology}
 In this section, we present our appraoch to Grouped Shapley Values (GSV). The approach is divided into multiple steps (1) definition of GSV for general cooperative games, (2) transfer of GSV to gain local explanations for machine learning models, (3) use of local explanations to gain global understanding, and finally (4) calculation of local explanations in polynomial time for tree structures.

\subsection{The Value of Predefined Coalitions in a Cooperative Game}
\label{sec:GSP}

As mentioned above, the classic Shapley value is defined to solve the fair distribution of resources within a cooperative game. A cooperative game is a tuple $(P,v)$, where $P=\{1,2, \dots, p\}$ is the finite set of players, and $v: 2^P \rightarrow \mathbb{R}$ is the value function. A subset of players $P$ is called a coalition, and the value function assigns a value to each coalition of players. The classic Shapley value now represents the average contribution of a player to all possible coalitions that the players in $P$ can form \citep{roth1988shapley}. For our approach, we extend the classic Shapley value formula by allowing the players to form predefined coalitions, before the game starts. Players in a predefined coalition will be evaluated together and will never be separated. To notate the predefined coalitions of players, we assume a partition of the set of players $\mathcal{C} = \{C_1, \dots, C_k\}$, where each $C_i$ is a nonempty subset of $P$ and represents a different predefined coalition. 

To obtain GSV, the classic definition of Shapley values is restricted to only average the contribution of the group $C_i$ to all possible subsets $S$ that can be built from predefined coalitions within $\mathcal{C}$. Therefore, the GSV $\varphi_{C_i}(v)$ for a predefined coalition $C_i$ depending on the value function $v$ can be defined as:

\begin{equation}\label{group_sv}
\varphi_{C_i}(v)=\sum_{S \subseteq \mathcal{C} \backslash\{C_i\}} \frac{|S| !\,(k-|S|-1) !\,}{k !\,}(v(\cup S \cup\{C_i\})-v(\cup S)),
\end{equation}
where $\cup S$ describes the union of all selected sets of $\mathcal{C}$ that are in $S$. 

We note that this definition allows the predefined coalition of players to have varying sizes, which is very useful in terms of feature importances, since natural groupings are mostly related to the origin of the feature and vary throughout most datasets. To understand GSV better, we can take a look at the fraction at the beginning of Equation \eqref{group_sv}. The fraction evolves from the original Shapley value definition that is made by permutating the set of players and averaging the difference when evaluating the value function over the coaltion with all players that precede a player $C_i$ in the given order with and without $C_i$ itself. The number of players preceding in our formula coincides with the number of players in the set $S$, having $S!$ possible orders. Similarly the players succeeding $C_i$ have $(k-|S|-1) !$ possible orderings. Normalizing with all possible $k!$ permutations results in the factor in Equation \eqref{group_sv}. Since the equation is a direct extension of the classical Shapley value, all desirable game-theoretic properties still hold (efficiency \eqref{eq:efficiency}, symmetry, dummy variable, and additivity). Most importantly, the efficiency axiom ensures that the Shapley value precisely distributes the gain produced by the coalition consisting of all players among all players.  
\begin{equation} \label{eq:efficiency}
	\text{\textbf{efficiency}}: \sum_{C_i \in \mathcal{C}} \varphi_{C_i}(v) = v(\mathcal{C}).
\end{equation}

Later, the axiom of efficiency translates into local explanations that always add up to explain the output value of the model for the explained data point. Exactly this important axiom of efficiency is not preserved in game-theoretic approaches to extend Shapley values on predefined coalitions that are proposed in \citep{grabisch1999axiomatic, marichal2007axiomatic, flores2019evaluating}. Therefore, these game-theoretic approaches are not applicable towards explainable machine learning, as the explanations would not add up to explain the model outcome. 

A naive solution to aggregate Shapley values within a group of players is to add individual Shapley values \citep{huber2022extreme}. To show that this option is not appropriate and produces counterintuitive results, we analyze a minimal example based on the classic illustrative glove game \citep{aumannShapley1974}. 

\textbf{\textit{The glove game example}}:
Within the glove game, we observe three players $P=\{1,2,3\}$ trying to complete a pair of gloves. The player $1$ and the player $2$ each have a left glove, while the player $3$ has a right glove. The value function $v(S)$ is evaluated with the value $1$, if the set $S$ contains a matching pair of gloves and with the value $0$ otherwise. 
Calculating the classic Shapley value, we obtain $\varphi_{1}(v)= \varphi_{2}(v) = \frac{1}{6}$ and $\varphi_{3}(v) = \frac{4}{6}$. The results follow our direct intuition that the player $3$ is the most important player in the game, since it is the only player who can complete a pair of gloves. Our observation changes when players $1$ and $2$ form a predefined coalition. This means $C_1 = \{1,2\}$ and $C_2 = \{3\}$. Both groups should be valued equally within the game, as having multiple left gloves does not increase the value function, and only a combination of $C_1$ and $C_2$ can build a pair of gloves. Using equation \eqref{group_sv} we observe that our definition of grouped Shapley values follows this intuition by valuing $\varphi_{C_1}(v) = \varphi_{C_1}(v) = \frac{1}{2}$, while when summing the initial values $\varphi_{1}(v) + \varphi_{2}(v) = \frac{2}{6}$ we would undervalue the coalition of players $1$ and $2$. Note that we can increase the gap between both approaches by adding more players who own a left-hand glove to the game.


%
\subsection{From Grouped Shapley Values to Local Explanations.}
\label{sec:ExplainByGSV}
To use GSV to better understand yield prediction, we define a game $(P,v)$ in the context of a machine learning model $M$ trained on a dataset $X \in \mathbb{R}^{n,m}$ with the following targets $y \in \mathbb{R}^{m}$. That is, the data set consists of n features $\{f_1, \dots, f_n\}$ and $m$ data points used to create the model. 

We obtain local explanations for a fixed data point $x \in X$ by interpreting each of the features as a player in a cooperative game, so $P = \{f_1, \dots f_n\}$. For a subset of features $S \subseteq P$, we want the value function $v(S)$ to represent the answer of the model $M$ to the data point $x$, assuming only the feature values of the data point from features in $S$ are known. 

For any predefined coalition $C_i \subseteq P$ and set $S \subseteq C \backslash\{C_i\}$, the difference between $v(\cup S)$ and $v(\cup S \cup \{C_i\})$ describes the change in the output of the model, assuming additional knowledge of the values of the features in $C_i$. The idea of the GSV is to average contributions over all possible combinations of other predefined coalitions $S$ and give an estimate of the impact of the specific values of the features in $C_i$ to the model output. 

\begin{figure}[ht]
    \centering
    \includegraphics[width=\textwidth]{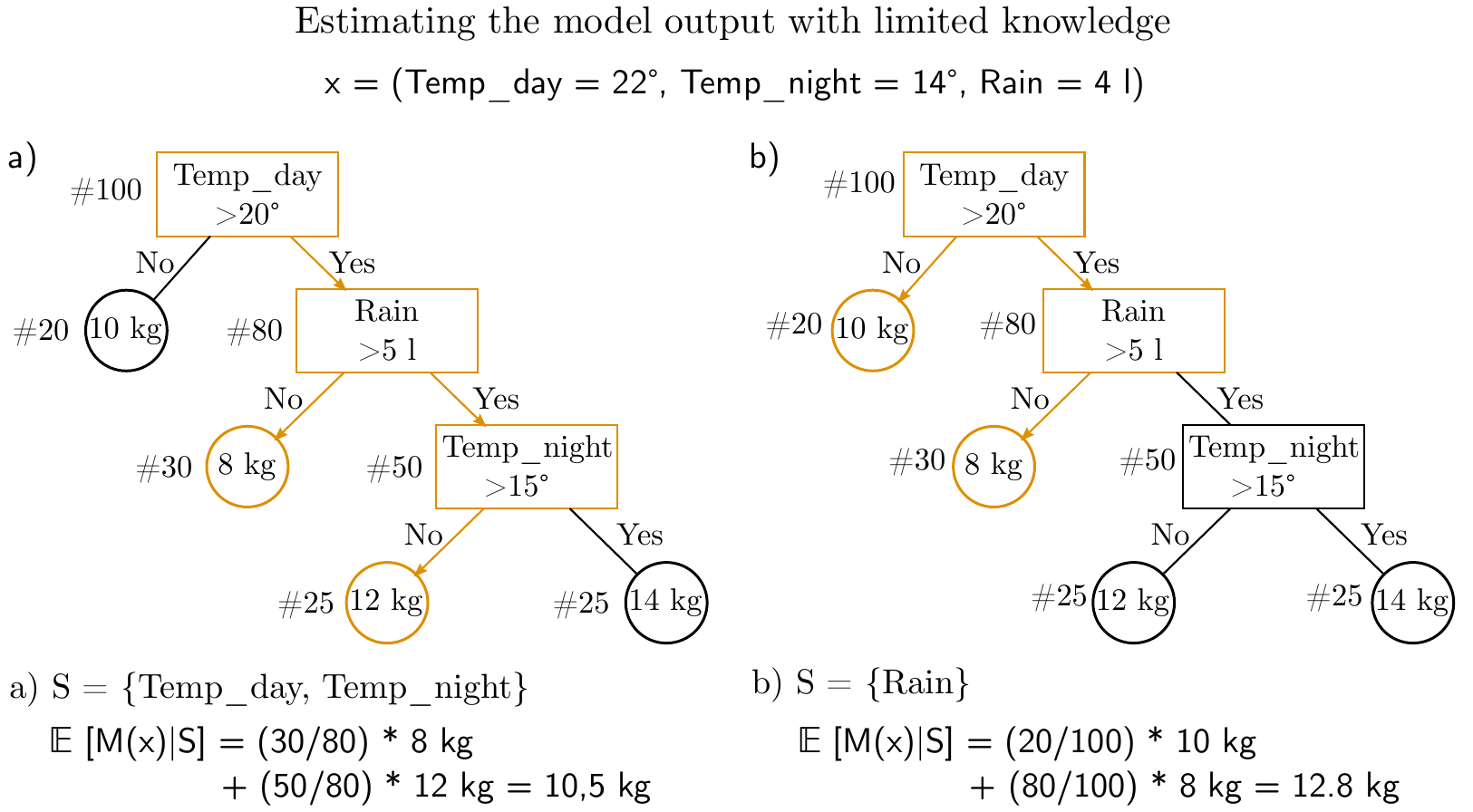}
    \caption{Two calculations of the the model output $\mathbb{E} [M(x)|S]$ for a fixed data point $x$, feature-subset $S$ and model $M$. The number of data points covered by each node is indicated by the number next to it. The same path ending in the prediction labeled ``8 kg'' is used to estimate the model answer of two different feature-subsets $S$.}
    \label{fig:tree_example}
\end{figure}

To give an estimation of the model answer based on limited access to features, we will calculate the expected answer of the model $M$ for a data point $x$, where only the features in $S$ are known: $\mathbb{E} [M(x)|S]$. To do so, we will take advantage of tree structures that have been shown to work well for yield prediction. To estimate the model's answer of a singular tree, we will traverse the tree as we would normally. If, while traversing, we find a feature $F$, that is not included in $S$ we estimate the average model answer from data points in our training data set that are similar to $x$ with regards to the set $S$. Similar data points are defined as points that induce the traversing of the tree equal to the data point $x$, for every feature in $S$. For the feature $F$ that is not in $S$ we calculate the weighted average of the model answer, according to the amount of similar data points that follow the 2 possible branches of the tree. We give exemplary calculations in Figure \ref{fig:tree_example} and put emphasis on handling the unknown feature ``Rain'' in part a) and ``Temp\_day'' in part b). The procedure described gives us a naive recursive algorithm to calculate the value function of Equation \eqref{group_sv} for any given data point and subset. By iterating over all the necessary subsets of features and building the sum, we can calculate the GSV. In Section \ref{sec:algo} we will describe a further procedure for calculating Equation \eqref{group_sv} in polynomial time.

\subsection{From Local Explanations to Global Understanding.}
\label{sec:localtoglobal}
The described approach for GSV is capable of obtaining local explanations for any yield prediction model, that is, predictions that can explain the impact of each feature for a specific data point. To continue, we want to have access to a global understanding of the model. Only then can we decide to drop features of low importance or analyze general patterns within the data that will help us to understand the yields. The low dimension of the GSV allows a clear representation of the importance of the features. The first step is to calculate the local GSV for a variety of data points. We can then utilize specialized \textbf{\textit{swarm plots}} for a joint representation to not only get an idea of the magnitude of impact each feature group possesses, but also the impact of high and low feature values on the final yield prediction. The information gained can then serve as baseline for further experiments to decide whether the machine learning task at hand is reliant on a specific group of features and can lead to decisions like, e.g. not buying a certain sensor or not using human resources to capture in-field information that often. 

We build the swarm plots as follows: After grouping the features, we decide on an aggregated value to represent the magnitude of the features in the group. For similar features, such as features captured from the same sensor in multiple timeframes, the mean value of all features serves this purpose. Since the number of groups is limited, we can even visualize multiple swarm plots in one figure, where each swarm represents a feature group. The x-axis is then used to show the GSV of the regarding group, indicating this feature group's impact on the prediction in the grand scheme. Finally, after normalizing the representative values for all groups, we can use the hue of the individual points to highlight how the high and low values of the feature groups influence the prediction of the model. Examples are shown in Figures \ref{fig:soybean} and \ref{fig:grapevine}.
Having the GSV calculated, it can help us to understand the impact of the group of features $C_i$ on our model for the specific data point $x$. Each GSV can be interpreted as the difference it made for our model, that the features in $C_i$ are valued within $x$ the way they are, in comparison to what the model would output, if these values were unknown. The higher the absolute value, the more impact is attributed to the feature group.

\subsection{Grouped Shapley Values on Tree Structures}
\label{sec:algo}
In general, the question of calculating the Shapley values is known to be NP-hard \citep{conitzer2004computing}. The usage of predefined a priori coalitions is capable of reducing the complexity of the task by reducing the number of summands in Equation \eqref{group_sv} compared to the classic Shapley value. We have already established how to estimate the value function $v(S)$ for a model $M$ and a fixed data point $x$ in Section \ref{sec:GSP} and how to use it to calculate Equation \eqref{group_sv}. Understanding the naive algorithm alters the way of formulating an algorithm capable of solving Equation \eqref{group_sv} in polynomial time. \citet{lundberg2020local} give a polynomial time algorithm for the calculation of the classic Shapley value feature importances on tree structures. On the basis of their work, we are able to formulate an algorithm to calculate the GSV feature importances in polynomial time. An abstract representation of the result can be seen in Algorithm \ref{alg:groupshap}, where we forego the exact weight update to achieve a more streamlined representation. An exact version of the algorithm can be found in the Appendix (Algorithm \ref{ap:alg:groupshap}).

\renewcommand{\thealgorithm}{1.\arabic{algorithm}}
\begin{algorithm}
\caption{Polynomial Group Shapley Values for Trees (Simplified)}\label{alg:groupshap}
\begin{algorithmic}[1]
\Procedure{GroupTreeShapleyValue} {$\textrm{Datapoint: }x, \textrm{ Model: }M, \textrm{ Coalitions: } \mathcal{C}$}
    \State $\varphi = \textrm{array of len(}  \mathcal{C} \textrm{) zeros}$ \Comment{Storage for GSV}
    \Procedure{Expand}{currentNode, path, weigths}
        \State path, weights = $\operatorname{Update\_weights}$(currentNode, path, weights) 
        \State $C_i$ = Group of currentNode.getFeature() \Comment{\parbox[t]{.35\linewidth}{Get the group of the node's split-feature}}
        \If{currentNode is an inner node}
            \If{a previous Feature F along the path is also in $C_i$}
                \State $\operatorname{UNDO}$ the $\operatorname{Update\_weights}$ for F \Comment{\parbox[t]{.36\linewidth}{Subset sizes remain the same}}
            \EndIf
            \State Child1 = Traverse further following the values of $x$
            \State Child2 = Traverse further the other child
            \State $\operatorname{Expand}$(Child1, path, weights)
            \State $\operatorname{Expand}$(Child2, path, weights)
        \EndIf
        \If{currentNode is a leaf}
            \For{node in path}
                \State $C_j$ = Group of node.getFeature()
                \State w\_frac = weights according to subsets along the path in regard to $C_j$
                \State w\_pos = weights of splits according to features in $C_j$
                \State w\_neg = weights of splits without information of features in $C_j$
                \State $\varphi$ [$j$] += $w\_frac * (w\_pos - w\_neg) *$ currentNode.getValue()
            \EndFor
        \EndIf
    
    \EndProcedure
    \State $\operatorname{EXPAND}(root, path = [], weights = [])$ \Comment{\parbox[t]{.4\linewidth}{Start at root node with empty path and no weights}}
    \State return $\varphi$
    \EndProcedure
    
\end{algorithmic}
\end{algorithm}

We have already described an intuition for calculating Equation \eqref{group_sv} by iterating over all possible subsets of the sum and estimating the answer of the value function. For each subset $S$ we have to traverse multiple paths of the tree, as highlighted in Figure \ref{fig:tree_example}, since every time we need to split at a feature not in $S$, we continue to build the weighted average over both possible following children. Similarly, we use each path of the model M in multiple calculations for the sets of Equation \eqref{group_sv}. In the examples in Figure \ref{fig:tree_example}, we see the path ending in ``8 kg'' traversed for two different subsets $S$. The main idea of the polynomial algorithm is to traverse the entire tree only once while keeping track of the contribution that each individual path has to all possible subsets. At the end of the traversal of each path, the grouped Shapley values are updated accordingly. The weights are determined by three different factors that need to be tracked and updated throughout the algorithm. First, the sizes of the possible sets $S$ as necessary for the factor in Equation \eqref{group_sv} (noted as w\_frac in Algorithm \ref{alg:groupshap}). Second, the fraction of training examples that follow the branches, as explained in the example in Figure \ref{fig:tree_example}. Third, for every group $C_i$ we want to calculate the GSV for, we need a sign depending on whether the path traversal assumes knowledge of the features in $C_i$ (positive) or not (negative). The weights are updated consecutively wihle traversing the tree. We focus on line 7 of the algorithm, where we need to check if the feature group was already represented during the path we traverse currently. If this is the case, the structure of subsets leading down to the path does not change, since all features within a group are treated as a unit.

The algorithm was first defined in \citep{lundberg2020local} for classic Shapley values. We altered the algorithm to calculate the grouped Shapley values. The correctness of our algorithm follows directly from the work of \citet{lundberg2020local}, since we only consider multiple features to be handled as if they were the same player within the equation. Since the check of occurring groups can be made via a lookup table, the run-time is still bounded by $O(TLD^2)$, with $T$ being the number of trees within the random forest. This is a factor since we need to execute the algorithm for every individual tree. $L$ refers to the maximum number of leaves within the trees and $D$ is the maximum depth of the trees. 

\section{Experimental Results}
\label{sec:exp}

To show how grouped Shapley values can be used for yield prediction, we will analyze models based on two different datasets for yield prediction. Each data set comes with different intricacies and challenges to explainability. The best insights about a data domain can be gained, when the models are well-fitted to the problem, since otherwise we would try to explain patterns that are not learned correctly. We build a regression forest for individual problems using e\textbf{X}treme \textbf{G}radient \textbf{Boost}ing (XGBoost) \citep{chen2016xgboost} and tune the models to be optimal using \textbf{T}ree-Structured \textbf{P}arzen \textbf{E}stimation (TPE) \citep{bergstra2013making}. XGBoost builds the regression tree ensemble from the training data by iteratively adding new trees to minimize the residual error of the training set, that is, the difference between the prediction made by the sum of all trees previously designed and the actual target output. For each dataset, we use 20\% of the training data for validation to apply TPE. The choice of testing sets depends on the data at hand, to be as close as possible to real-world scenarios.

\subsection{Soybean Yield Prediction based on Remote Sensing Data}
\begin{figure}
    \centering
    \includegraphics[width=\textwidth]{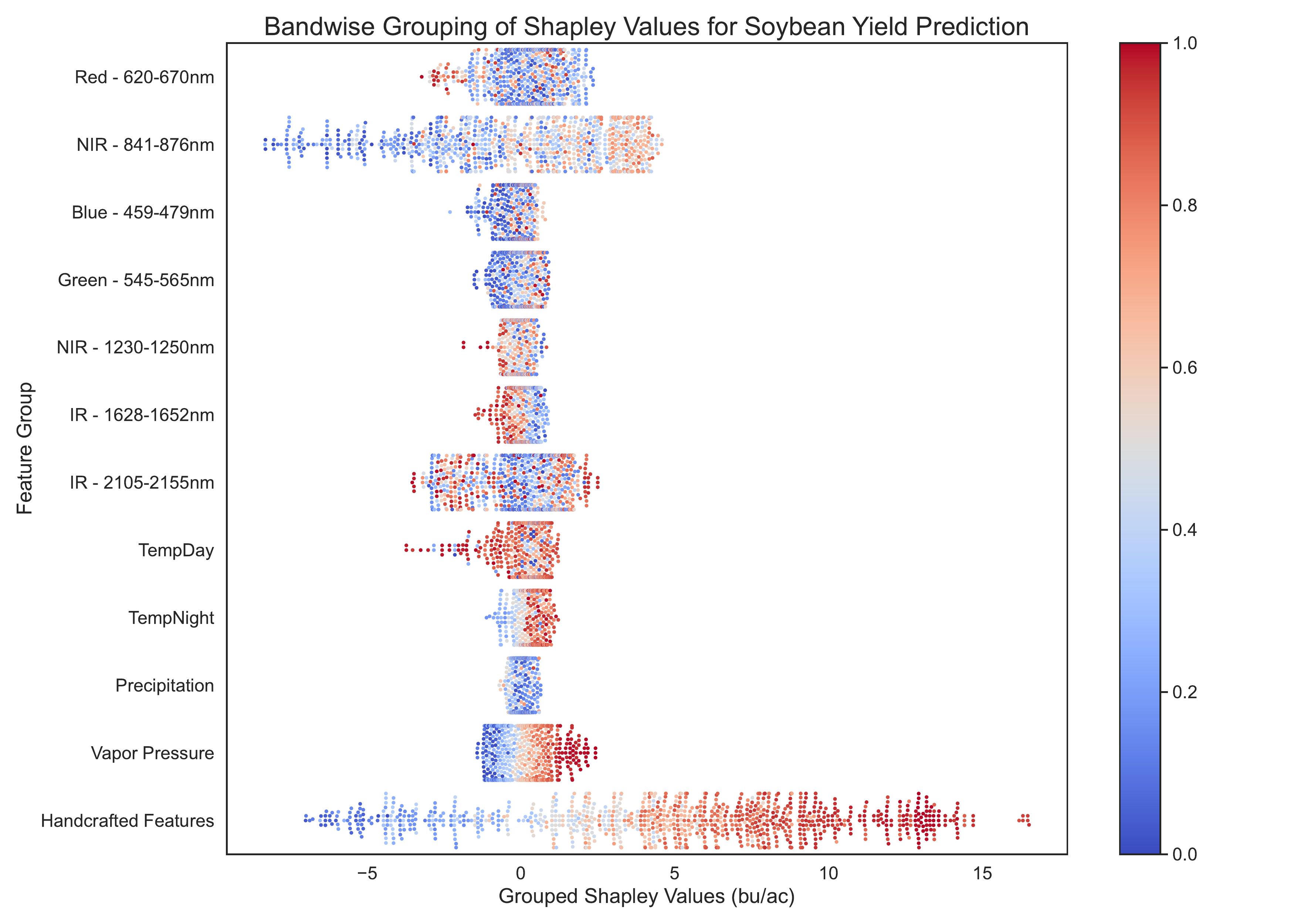}
    \caption{GSV feature importances for soybean yield prediction. Every dot is a local explanation of a soybean yield prediction in the U.S. The first 11 plots represent a group of 102 features each calculated from different remote sensing data. The bottom plot shows a group of 7 additional handcrafted features. The color represents an averaged and normalized representation of the grouped features values.}
    \label{fig:soybean}
\end{figure}

The first model is trained to predict county-level soybean yields in the United States. The real-world yield data are taken from the U.S. Department of Agriculture \citep{USDA}. Grouped Shapley values are necessary here, since the input data consist of 1131 features. The features are taken from remote sensing images and originate from 11 different bands captured in 8 day composites at 34 different time steps. The first 7 bands are from the MODIS satellite and differ within the wavelength depicted. The temperature bands are also MODIS products and capture the average temperature during the day and at night \citep{vermote2015mod09a1}. The precipitation and vapor pressure bands are taken from the Daymet V4 satellite \citep{thornton2016daymet}. Following \citep{huber2022extreme}, each band at each time is compressed to three values, that is, the mean, a 20\% quantile and an 80\% quantile. 
Figure \ref{fig:grapevine} shows the swarm plots resulting from the grouping of features with respect to their spectral bands. 
We note that the sum of all GSV of a local explanation will result in the model output for the specific data point. Therefore, each grouped Shapley value describes how knowing the data point's values for the features within the groups impacted the model output, in comparison to the absence of these values. If we look, for example, at the values on the far right within the swarm plot for the handcrafted features in Figure \ref{fig:soybean}, we know that there are data points that we would expect to have ca. 15 bu/ac lower predicted yield, when the handcrafted features would be unknown. Furthermore, we know that this behavior is caused by the high overall values of the feature group, since the respective data points are colored in red. When iterating through the swarm plots, we see interesting and insightful patterns in most of them. The following insights can be gained from the corresponding swarm plots:

\textbf{Red - 620-670 nm and NIR - 841-876:} These two feature groups are components of the well-known Normalized Difference Vegetation Index (NDVI) \citep{quarmby1993use}, which is historically used to summarize remote sensing images to predict yields. The NDVI is calculated by dividing the difference between the NIR and the Red band by their sum. This means that a high NIR will increase the NDVI and is correlated with a higher yield, while the opposite holds for the red band. This coincides with the plots in Figure \ref{fig:soybean}, as can be seen by the reverse order of the red and blue points in the two plots. For the red band, the blue dots are mostly on the positive side of the plot, meaning that a low value for this band coincides with an increasing yield prediction. For the NIR band, we observe the opposite. The dots colored blue make the biggest negative impact on the model of all feature groups, by reducing the model's prediction by more than 5 bu/ac.

\textbf{Blue - 459-479 nm and Green - 545-564 nm:} Both bands have historically not been used for yield prediction and also have little impact on our prediction models, as there are no dots within the plots that show a high GSV.

\textbf{NIR - 1230-1250 nm and IR - 1628-1652 nm}:
These feature groups show lower impacts, indicated by very narrow swarm plots. But for both bands, we see a tendency that higher values coincide with lower predictions since the negative impacts on the yield predictions are all recorded for red-colored dots.

\textbf{IR - 2105-2155 nm:} 
This feature group shows a higher impact, indicated by a larger swarm plot. Since we find red and blue dots at both ends of the spectrum, we cannot derive a pattern or interpret the influence of higher or lower values. This means that the model's interpretation of this feature group is highly influenced by other feature groups around, but still important to derive the final prediction.

\textbf{TempDay and TempNight:}
The temperature at day shows a larger swarm plot than the temperature at night and therefore is more influential on the model output. At the same time, the dots within the night temperature are clearly sorted from blue to red and indicate that a higher temperature at night coincides with a higher yield prediction.

\textbf{Precipitation and Vapor Pressure:}
The precipitation group shows almost no impact on the model output, as it is the narrowest swarm plot within Figure \ref{fig:tree_example}. The vapor pressure group shows a small impact on the yield prediction and a very clear indication that a higher vapor pressure should lead to slightly higher yields, because the dots are completely in order from blue to red. Interestingly, both feature groups are not available worldwide, as they are specifically captured within the United States. The relatively low impact on the prediction model encourages soybean yield prediction experiments in other regions of the world, although this information would not be available.

\textbf{Handcrafted Features:}
This is the most influential group of features, capable of altering the prediction of yield by more than 15 bu/ac. The impact is very high, as it includes the average yield of the county represented by the data point over the foregoing years. That is, a county with traditionally higher yields in the past will obtain a higher yield prediction by our model, thus inducing spatial context to the modeling.

\subsection{Grapevine Yield Prediction based on Phenological Data}

\begin{figure}[ht]
    \centering
    \includegraphics[width=\textwidth]{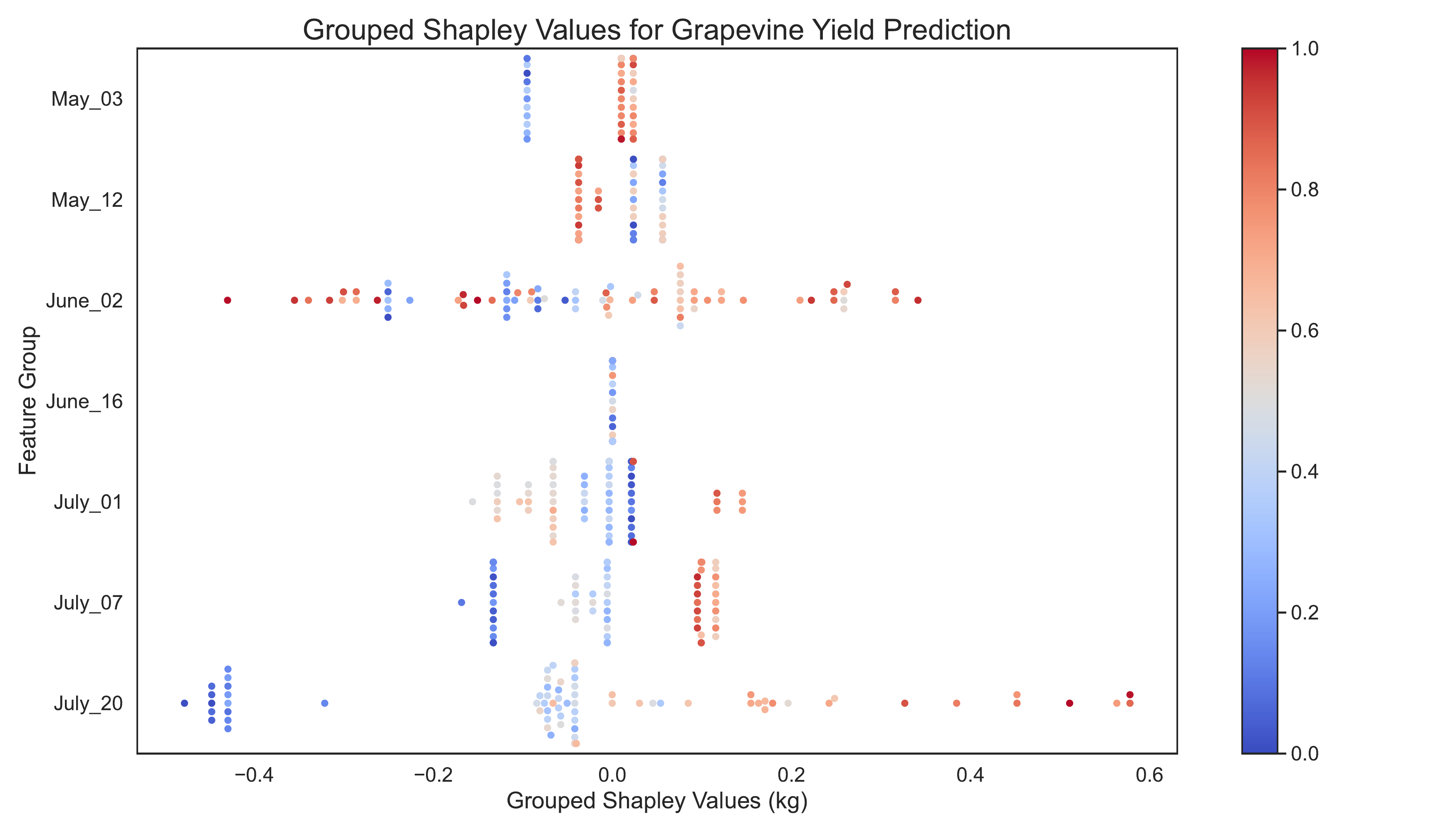}
    \caption{GSV feature importances for Grapevine yield prediction. Each plot represents a group of features that are captured at the same point in time. The color represents an averaged and normalized representation of the grouped features values.}
    \label{fig:grapevine}
\end{figure}

The second model, which we use to highlight the ability to understand yield prediction models through GSV feature importance, predicts grapevine yields. The data are directly obtained from the Julius Kühn-Institute, an instiute for grape vine breeding in germany. The input features are measured by manual plant appraisal in the year 2021 and the data consist of 461 grapevines stemming from 11 different grape varieties, with manual weighted yields, directly after harvest. The 11 different grape varieties include established varieties already used in commercial practice and experimental varieties, which are not guaranteed to meet the high conditions required to become commercially viable. Data are captured at 7 different timestamps and consist of different phenological information, such as the number of grapevine shoots, the number of inflorescences, and the number of clusters of grapes. GSV can help to identify important and nonimportant time-stamps to obtain meaningful yield predictions. Since the data are captured manually, omitting some time stamps for data retrieval could save time and money. We divide the data into test and training subsets by randomly sampling 20\% of the data for testing. The GSV for the test data can be seen in Figure \ref{fig:grapevine}. Analyzing the plots one by one, we see the following patterns:

\textbf{May\_03 and May\_12:} Both times of data acquisition are early in the grapevine growth cycle. The dots for both plots are close to the center, which means that the impacts of both feature groups are rather low. However, we still observe some patterns. For the feature group May\_03 we see that all the blue dots have a negative GSV and therefore a negative impact on the predictions. Within the feature group, we count the number of shoots. We can deduce that too few shoots in the early stages of growth will harm the expected yield. 

\textbf{June\_02:} This is one of the groups showing one of the largest impacts on our model, as can be seen from the GSV in a range between -0.4 kg and 0.4 kg. The high importance of this at such an early stage of the growth cycle may seem surprising at first. Investigating the issue, with the help of knowledge from domain experts, helps us explain why this feature group is important. For grapevines in our study region, it is known that grapes build their inflorescences up to the beginning of June. Since this phenological feature is counted within this study group, it helps explain its high importance. Regarding the distribution of values within this feature group, we see that the red dots are distributed at both ends of the plot, whereas the blue dots are mostly in the negative area. We can interpret this to mean that a high value of the phenological features of the group is a necessary but not a sufficient criterion for a high yield prediction.

\textbf{June\_16 and July\_01:} The feature group of features captured in June\_16 is of no importance to our model, since all GSV are zero-valued. The group of features captured on July\_01 shows an interesting pattern. We see that the dots with negative GSV are all colored very lightly, meaning that the values are close to the average of the regarding feature group. Slightly better yields can be expected for low feature values, as indicated by the cluster of blue dots, and the best yield is expected for high values of the feature group, as indicated by the cluster of red dots on the right. 

\textbf{July\_20} This feature group contains all the features captured on July 20th, and therefore the captured information is very close to harvest. Not only do we see GSV from -0.4 to 0.6, we also see a clear pattern with the dots being ordered from blue to red. The information captured includes phenological information that counts the number of clusters of grapes, which directly correlates with the expected yield. 

\section{Conclusions}
\label{sec:Conclusion}

In this work, we propose a novel approach to explain machine learning models in the context of yield prediction. First, we examine a notion of Shapley values for predefined coalitions in a cooperative game that is capable of preserving desirable mathematical properties of the classical Shapley value. Foremost, the axiom of efficiency that states, when transferred to the context of feature importances, that the sum of all Shapley values of a specific explanation will add up to show the model's output. We call this Grouped Shapley Values (GSV). Second, we leverage the visualization of many local explanations, to achieve global understanding of our model. This is done by introducing swarm plots that not only show the GSV but also give a colored indication of the feature's aggregated values, giving further information to analyze. Third, we solve the NP-hard problem of calculating the GSV in polynomial time for tree structures. On the one hand, tree structures show state-of-the-art results for many yield prediction scenarios, on the other hand, we can calculate the GSV by traversing a tree once and keeping track of the according addition to the GSV, that comes from each singular path. And fourth, we give an exemplary analysis on two real-world yield prediction tasks, showing the worth that GSV can add to yield prediction. We can not only learn from the predictions, for example, about conditions that favor high yields, but we can also raise trust in our models. This is achieved by showing that the important features are in line with the features that domain experts would consider for yield prediction. We conclude that our approach will be helpful in creating synergy between computer scientists and domain experts. 
We acknowledge that our approach relies on a natural grouping of features that cannot be determined in every scenario, but is common for yield prediction tasks, as multiple features are often derived from the same sensor or are captured at the same time frames. Also, our visualization works best, if the feature groups values can be represented meaningfully by an aggregated value. Lastly, since fast calculation of the GSV is only enabled for tree structures, it remains an open problem how to handle explanations for models with high-dimensional input data, that do not rely on tree structures. For future work, we will extend our approach to different datasets, and we will address the problem that occurs when no natural grouping of features is accessible. Furthermore, we will experiment on using the calculated feature importances to select features and create lightweight models for yield prediction.

\section*{\begin{normalsize}Acknowledgements\end{normalsize}}
\noindent
This work was partially done within the project ``Artificial Intelligence for innovative Yield Prediction of Grapevine” (KI-iREPro). The project is supported by funds of the Federal Ministry of Food and Agriculture (BMEL) based on a decision of the Parliament of the Federal Republic of Germany. The Federal Office for Agriculture and Food (BLE) provides coordinating support for artificial intelligence (AI) in agriculture as funding organisation, grant number FKZ 28DK128B20. 
\noindent
We thank Timm Haucke for proofreading our manuscript. 
\bibliography{main.bib}

\appendix
\section{Appendix A: Algorithms}
\label{sec:appendix}
Based on the work of \citep{lundberg2020local} we present Algorithm \ref{ap:alg:groupshap} as a precise description of Algorithm \ref{alg:groupshap}. Regarding the notion within the algorithm, the model M is always a tree represented by a list of nodes $v$, left children $a$, right children $b$, thresholds $t$, cover $r$, and features $d$. The cover is the fraction of training data that are split by the individual nodes. The variable $m$ is used to store the path of the unique feature groups along the path, represented by one feature per group. Together with the path $m$, we store four attributes. (1) The feature index $d$, (2) the fraction of paths, where this group is not in the set $S$ that flow through the branch $z$, (3) the fraction of paths, where this feature group is in the set $S$ - $o$, and finally (4) the weight $w$, which keeps track of the weights in front of Equation \eqref{group_sv}. Within the algorithm, we access arrays via dot notation and $m.d$ represents the whole vector of features that are traversed so far. Lastly, the values $p_z$ and $p_o$ track the fraction of added contributions, depending on the current feature as represented in the subsets.

\begin{algorithm}
\scriptsize
\caption{Polynomial Group Shapley Values for Trees (Detailed)}\label{ap:alg:groupshap}
\begin{algorithmic}[1]
\Procedure{GroupTreeShapleyValue} {$x, tree = \{v,a,b,t,r,d\}, \mathcal{C}$}
    \State $\varphi = \textrm{array of } \mathcal{C} \textrm{ zeros}$ \Comment{Storage for GSV}
    \Procedure{Expand}{$j,m,p_z, p_o, p_i$}
        \State $m =  \operatorname{Weight\_update} (m, p_z, p_o, p_i$) \Comment{\parbox[t]{.4\linewidth}{Update m and all fractions to incorporate the growing amount of features included in the path}}
        \If{$v_j \neq internal$}  \Comment{$v_j$ is a leaf node}
            \For{$i \leftarrow$ to len($m$) } \Comment{Traverse path backwards}
                \State $w= sum( \operatorname{Unwind} ((m,i).w$) \Comment{\parbox[t]{.4\linewidth}{Undo $\operatorname{Update\_weight}$ to access all features along the path}}
                \State $C =\textrm{ group of }d_j$ \Comment{Find group of the current feature}
                \State $\varphi_{C}=\varphi_{C}+w\left(m_i \cdot o-m_i \cdot z\right) v_j$ \Comment{Add contribution to SHAP values}
            \EndFor
        \Else \Comment{$v_j$ is an internal node}
            \State $h, c=\left(a_j, b_j\right)$ if $x_{d_j} \leq t_j$ else $\left(b_j, a_j\right)$ \Comment{\parbox[t]{.35\linewidth}{check which path the feature values of x dictate}}
            \State $i_z=i_o=1$
            \State $k=\operatorname{FINDFIRSTGROUP}\left(m . d, d_j\right)$ \Comment{Check for group of $d_j$}
            \If{k $\neq$ nothing} \Comment{Undo split if the group is already represented}
                \State $i_z, i_o = (m_k.z, m_k.o)$
                \State $m = \operatorname{Unwind} (m,k)$
            \EndIf
        \State $\operatorname{Expand}\left(h, m, i_z r_h / r_j, i_o, d_j\right)$ \Comment{Recursive call for both children}
        \State $\operatorname{Expand}\left(c, m, i_z r_c / r_j, 0, d_j\right)$
        \EndIf
    \EndProcedure
        \Procedure{Weight\_update}{$m, p_z, p_o, p_i$}
        \State $l,m = len(m), copy(m)$
        
        \State $\textrm{subsetsize} = 1 \textrm{ if } l = 0, \textrm{ else } \textrm{subsetsize} = 0$ \Comment{Check if this is the first call}
        \State $ m_{l+1} \cdot(d, z, o, w)=\left(p_i, p_z, p_o, subsetsize\right) $ 
        \For{$i \leftarrow \text{ to } 1$}
            \State $m_{i+1} \cdot w=m_{i+1} \cdot w+p_o \cdot m_i \cdot w \cdot(z / l)$ \Comment{Fraction for bigger subsets}
            \State $m_i \cdot w=p_z \cdot m_i \cdot w \cdot(l-i) / l$ \Comment{Fraction for same size subsets}
        \EndFor
        \State return $m$
    \EndProcedure
    \Procedure{Unwind}{m,i}
        \State $l,n,m = len(m), m_l.w, copy(m_{1 \dots l-1})$
        \For{$i \leftarrow l-1 \text{ to } 1$} \Comment{Run the path backwards}
            \If{$m_i.o \neq 0$}
                \State $t=m_j \cdot w$ \Comment{Undo the calculations within $\operatorname{Weight\_update}$}
                \State $m_j \cdot w=n \cdot l /\left(j \cdot m_i \cdot o\right)$
                \State $n=t-m_j \cdot w \cdot m_i \cdot z \cdot(l-j) / l$
            \Else
                \State $m_j \cdot w=\left(m_j \cdot w \cdot l\right) /\left(m_i \cdot z(l-j)\right)$
            \EndIf
        \EndFor
        \For{$j \leftarrow i \text{ to }l-1$} \Comment{Update weights}
            \State $m_j \cdot(d, z, o)=m_{j+1} \cdot(d, z, o)$
        \EndFor
        \State return $m$
    \EndProcedure
    \State $\operatorname{Expand}(1, [], 1,1,0)$ \Comment{Start at root node}
    \State return $\varphi$
    \EndProcedure

\end{algorithmic}
\end{algorithm}

%
%

\end{document}